\documentclass[10pt,twocolumn,letterpaper]{article}

\usepackage{cvpr}              

\usepackage{graphicx}
\usepackage{amsmath}
\usepackage{amssymb}
\usepackage{booktabs}

\usepackage{graphicx}
\usepackage[ruled,vlined]{algorithm2e}
\usepackage{xcolor}         

\usepackage[accsupp]{axessibility}  

\definecolor{cobalt}{rgb}{0.0, 0.18, 0.39}

\usepackage{amsmath,amsfonts,bm}

\def\eqref#1{equation~\ref{#1}}

\def\Algref#1{Algorithm~\ref{#1}}

\def\1{\bm{1}}

\def\eps{{\epsilon}}

\DeclareMathAlphabet{\mathsfit}{\encodingdefault}{\sfdefault}{m}{sl}
\SetMathAlphabet{\mathsfit}{bold}{\encodingdefault}{\sfdefault}{bx}{n}

\newcommand{\E}{\mathbb{E}}

\usepackage[pagebackref,breaklinks,colorlinks]{hyperref}

\usepackage[capitalize]{cleveref}
\crefname{section}{Sec.}{Secs.}
\Crefname{section}{Section}{Sections}
\Crefname{table}{Table}{Tables}
\crefname{table}{Tab.}{Tabs.}

\def\cvprPaperID{891} 
\def\confName{CVPR}
\def\confYear{2022}

\begin{document}

\title{Simulated Adversarial Testing of Face Recognition Models}

\author{Nataniel Ruiz\\
Boston University\\
{\tt\small nruiz9@bu.edu}
\and
Adam Kortylewski\\
Johns Hopkins University\\
{\tt\small akortyl1@jhu.edu}
\and
Weichao Qiu\\
Johns Hopkins University\\
{\tt\small qiuwch@gmail.com}
\and
Cihang Xie\\
UC Santa Cruz\\
{\tt\small cixie@ucsc.edu}
\and
Sarah Adel Bargal\\
Boston University\\
{\tt\small sbargal@bu.edu}
\and
Alan Yuille\footnotemark[1]\\
Johns Hopkins University\\
{\tt\small ayuille1@jhu.edu}
\and
Stan Sclaroff\footnotemark[1]\\
Boston University\\
{\tt\small sclaroff@bu.edu}
\and
}

\maketitle

\begin{abstract}
     Most machine learning models are validated and tested on fixed datasets. This can give an incomplete picture of the capabilities and weaknesses of the model. Such weaknesses can be revealed at test time in the real world. The risks involved in such failures can be loss of profits, loss of time or even loss of life in certain critical applications. In order to alleviate this issue, simulators can be controlled in a fine-grained manner using interpretable parameters to explore the \textit{semantic image manifold}. In this work, we propose a framework for learning how to test machine learning algorithms using simulators in an adversarial manner in order to find weaknesses in the model before deploying it in critical scenarios. We apply this method in a face recognition setup. We show that certain weaknesses of models trained on real data can be discovered using simulated samples. Using our proposed method, we can find adversarial synthetic faces that fool contemporary face recognition models. This demonstrates the fact that these models have weaknesses that are not measured by commonly used validation datasets. We hypothesize that this type of adversarial examples are not isolated, but usually lie in connected spaces in the latent space of the simulator. We present a method to find these \textit{adversarial regions} as opposed to the typical adversarial points found in the adversarial example literature.
\end{abstract}

\renewcommand{\thefootnote}{\fnsymbol{footnote}}
\footnotetext[1]{Equal senior contribution.}

\section{Introduction}
\label{sec:intro}

Evaluating a machine learning model can have many pitfalls. Ideally, we would like to know (1) when the model will fail (2) in which way it will fail and (3) how badly it will fail. In other words, we would like to be able to accurately estimate the model's risk on the true test data distribution as well as know what specific factors induce the model to failure. We would like to know how these failures will manifest themselves. For example, whether a face verification model will generate a false-positive or false-negative error. And finally, when this failure happens, we would like to know how confident was the incorrect decision by the model. Testing models is no longer a purely academic endeavour~\cite{yuille2021deep}, with many high profile bad societal consequences being revealed in recent years due to insufficient testing particularly with respect to racial and gender bias in face analysis systems~\cite{buolamwini2018gender, Garcia_2019, grother2019face}.

These three desiderata are very hard to achieve in practice. There are major philosophical and theoretical obstacles to achieve perfect knowledge of model failures a priori. Nevertheless, partial knowledge of model weaknesses and predictions of model failures are possible. Yet, there are still major hurdles that stand in our way.

One such hurdle is the fact that \textit{testing data is limited}, due to the fact that it is expensive to gather and label. It is not uncommon for a model to perform well on an assigned test set and fail to generalize to specific obscure examples when it is deployed. A second important hurdle is the fact that \textit{testing data is unruly}. There are latent factors that generate the testing data, which are hard to control or even to fully understand. For example, a known factor that is hard to control is the lighting of a scene. Most datasets have been captured without controlling for this variable, and thus present an insufficient amount of variability in this respect. Testing a model in one environment could yield perfect performance, yet fail on an environment with more lighting variability. Even if a test dataset with carefully controlled lighting were assembled, the dataset would be very expensive and time-consuming to collect and there is no guarantee that the full variability would be explored.

A way to tackle these problems is to use simulators to generate test data. Such an approach can cheaply generate a large quantity of data spanning a large spectrum. Also, simulators are fully controllable and the generative parameters are known. This allows for careful exploration of situations where models fail. This includes the possibility to find intepretable factors that generate failures, to study the way these failures manifest themselves (is the model classifying a cat as a jaguar when there is green in the background?) and to examine the degrees of certainty of the model in these failure modes.

When simulating test data, we have full control over simulator parameters. Thus, we are able to explore the manifold generated by the simulator in the space of the simulator parameters. We call this manifold the \textit{semantic image manifold}, in contrast to the \textit{adversarial image manifold} that is explored in the traditional adversarial attack literature. A random exploration of this manifold is both inefficient and not the most informative approach. In this work we propose to \textbf{\textit{test machine learning models using simulation in an adversarial manner}} by finding simulator parameters that generate samples that fool the model. We are inspired by the literature on adversarial examples that fool machine learning models, yet in contrast to this body of work, the adversarial examples that our simulator generates are \textit{semantically realistic} in the sense that we are not adding low magnitude noise to an image in order to fool the model but finding semantically sensible image configurations that generate model failure. In this way, we are not investigating the well-known weakness of gradient-based models to unrealistic targeted noise but to plausible scenes that might be rare, yet mislead the model. We present a method that finds adversarial samples efficiently using a continuous policy that searches the high-dimensional space of possibilities.

A limitation of this type of work is that, in general there exists domain shift between the distribution described by the simulator and the real world distribution~\cite{Ganin:2016:DTN:2946645.2946704,Chen_2017_ICCV,tzeng2017adversarial, Tsai_2018_CVPR, hoffman2018cycada, peng2019moment}. Nevertheless, in our work we are able to show that in some situations, real model weaknesses can be found using simulated data. This gives credence to the hypothesis that, even though there is domain shift, simulated samples can be informative. Also, simulators are rapidly improving in terms of realism~\cite{ruiz2020morphgan, mildenhall2020nerf, gafni2020dynamic, li2021neural}. This allows for greater opportunities to use these ideas in the future as simulated and real data distributions become more and more aligned.

We hypothesize that these adversarial examples are not isolated points in space, but instead are regions of this manifold. In prior work on traditional adversarial examples, optimization procedures find adversarial samples that are points in image space~\cite{szegedy2013intriguing, papernot2017practical, carlini2017towards, explaining_adv, madry2018towards, salman2020unadversarial}. In contrast to this body of work we propose a method to find these \textit{adversarial regions} instead. This is valuable because ideally we would like to be able to fully describe the machine learning model's \textit{regions of reliability}, where model predictions will tend to be correct. With this knowledge a user would be able to avoid performing inference on a model outside of its scope in order to minimize failures. 

Contributions of this work are three-fold. We summarize them as follows:
\begin{itemize}
    \item We show that weaknesses of models trained on real data can be discovered using simulated samples. We perform experiments on face recognition networks showing that we can diagnose the weakness of a model trained on biased data.
    \item We present a method to find adversarial simulated samples in the \textit{semantic image manifold} by finding adversarial simulator parameters that generate such samples. We present experiments on contemporary face recognition networks showing that we can efficiently find faces that are incorrectly recognized by the network.
    \item We present a method to find \textit{regions} that are adversarial, in order to locate danger zones where a model's predictions are more liable to be incorrect. To the best of our knowledge, we are the first to explore the existence of these adversarial regions in the interpretable latent space of a simulator.
\end{itemize}

\begin{figure*}[t]
\centering
\includegraphics[clip,width=0.65\textwidth]{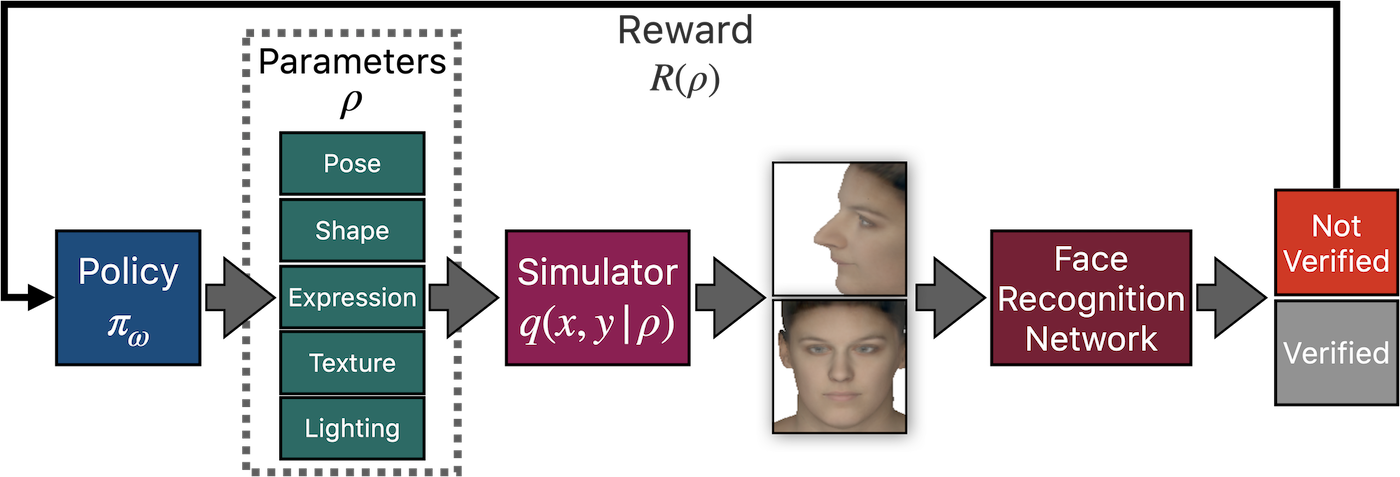}
\caption[]{Our method applied to the face verification scenario. The simulator is conditioned on parameters generated by the policy. An image pair of the same identity is generated. Face verification is run on this image pair using the face recognition network that is to be diagnosed. A reward is computed based on the correct or incorrect prediction of the network and policy parameters are updated accordingly.
\label{fig:pipeline}}
\vspace{-20px}
\end{figure*}

\section{A Framework for Simulated Adversarial Testing}
\label{sec:method}

Here we formalize adversarial testing using a simulator. We postulate some assumptions on the data generation process in the real and simulator world. Then we give the risks for a machine learning model and the mathematical formulation to find adversarial parameters that yield samples that fool machine learning models. We then present some parallels between our scenario and the literature on learning across domains. Finally, we describe our proposed algorithm to find such adversarial simulator parameters and adversarial samples.

Let us assume the real world data $(x,y)$ (where $x$ is the data and $y$ is the label) is generated by the distribution $p(x,y|\psi)$ where $\psi$ is a latent variable that causally controls the data generation process. For example, $\psi$ includes the object type in the image and the angle of view of such an object, as well as all other parameters that generate the scene and image. The risk for a discriminative model $f$ is:
\begin{equation}
\mathbb{E}_{\psi \sim a}[\mathbb{E}_{(x,y) \sim p(x,y|\psi)}[L(f(x),y)]],
\end{equation}
where $a$ is the distribution of $\psi$ and $L$ is the loss.
We can search for $\psi^*$ that maximizes this risk:
\begin{equation}
\max_{\psi \in A}[\mathbb{E}_{(x,y) \sim p(x,y|\psi)}[L(f(x),y)]]
\end{equation}
where $A$ is the set of all possible $\psi$.
Let us assume that we have $\psi=(\psi_u,\psi_k)$, a decomposition of $\psi$ into two latent variables $\psi_u$ and $\psi_k$. Furthermore, let us assume that $\psi_u$ controls for unknown features of the image, and $\psi_k$ controls for known features of the image such as the camera pose, or the object position with respect to the camera. We can write the average risk as:
\begin{equation}
\mathbb{E}_{\psi_u \sim a}[\mathbb{E}_{\psi_k \sim b}[\mathbb{E}_{(x,y) \sim p(x,y|\psi_u, \psi_k)}[L(f(x),y)]]],
\end{equation}
where $b$ is the distribution of $\psi_k$.
In most scenarios, we do not have access to the real data distribution $p$ and cannot sample from it at will. Additionally, it is very difficult to control the known latent variable $\psi_k$ when generating data, and we do not even know what factors are hidden in the variable $\psi_u$, much less how to control it. Using simulated data we are able to fully control the generative process. 

A simulator samples data $(x,y) \sim q(x,y|\rho)$, where $q$ is the simulated data distribution and we have complete knowledge over the latent variable $\rho$. We are able to search for adversarial examples and compute estimates of the mean and worst-case risks using this simulator. For example, the parameter $\rho^*$ that maximizes the risk is written as follows:
\begin{equation}
\max_{\rho \in C}[\mathbb{E}_{(x,y) \sim q(x,y|\rho)}[L(f(x),y)]]
\end{equation}
where $C$ is the set of all possible $\rho$. We can find $\hat{\rho^*}$, an estimate of $\rho^*$, by sampling (albeit inefficiently). In our case we are working in a less restrictive scenario since we do not try to find the global maximum $\rho^*$, instead we try to find any $\rho$ where $\mathbb{E}_{(x,y) \sim q(x,y|\rho)}[L(f(x),y)]$ is above the misclassification threshold.

If we assume that the distributions $p$ and $q$ are similar enough we can use the knowledge gathered in simulation to understand the possibilities of failure in the real world. Essentially, this is a different kind of domain shift problem. In a traditional setting of transfer learning between domains, we are concerned about minimizing the risk on a target domain by training on a source domain. In the binary classification case, let us define a domain as a pair consisting of a distribution $p$ on inputs $\mathcal{X}$ and a labeling function $g_p: \mathcal{X} \rightarrow [0,1]$. We consider the \textit{real domain} and the \textit{simulated domain} denoted by $(p, g_p)$ and $(q, g_q)$ respectively.

We also introduce a \textit{hypothesis} that is a function $h:\mathcal{X} \rightarrow \{0,1\}$. We can write the risk of this hypothesis on $p$ as:
\begin{equation}
\epsilon_p(h,g_p) = \mathbb{E}_{x \sim p}[|h(x) - g_p(x)|]
\end{equation}

In traditional domain adaptation from simulation to reality, we seek to learn on distribution $q$ and generalize to distribution $p$. We want to find a hypothesis that minimizes the risk on the target real world distribution $\epsilon_p(h,g_p)$ by training on samples from $q$.

In our setting, we do not train on synthetic samples. Instead we want to find a relationship between testing a hypothesis $h$ on samples from distribution $q$ and testing $h$ on samples from $p$. There exist bound results for the risks $\epsilon_p(h,g_p)$ and $\epsilon_q(h,g_q)$ in the work of Ben-David \textit{et al.}~\cite{ben2010theory}:
\begin{multline}
\epsilon_p(h,g_p) < \epsilon_q(h,g_q) + d_1(q,p) + \\ \min\{\mathbb{E}_p[|g_q(x)-g_p(x)|], \mathbb{E}_q[|g_q(x)-g_p(x)|]\},
\end{multline}
where $d_1$ is the variation divergence. The second term of the right hand side quantifies the difference between distributions $q$ and $p$, and the third term of the right hand side is the difference between the labeling functions across domains, which is expected to be small.

Since this bound characterizes the cross-domain generalization error and $\epsilon_q(h,g_q)$ will usually be minimized by the learning algorithm, it is useful for studying transfer learning between domains. There are some differences in our scenario since for us $h$ is a fixed function that has been trained on the target domain and we would like to talk about individual examples instead of overall risk over distributions. Also, the bound is proven for a binary classification problem, whereas our target scenario can be multi-class classification or regression.

Assume there exists a mapping $\tau: C \rightarrow A$, that maps the simulated latent variables to real latent variables $\psi = \tau(\rho)$. In order for adversarial examples in the simulator domain to be informative in the real domain, we want to have a simulator such that:
\begin{equation}
\mathbb{P}_{(x_s,y_s) \sim q, (x_r, y_r) \sim p}[|L(x_s, y_s) - L(x_r, y_r)| < \eps] > \theta.
\end{equation}
We denote $p(x_r,y_r|\tau(\rho))$ as $p$ and $q(x_s,y_s|\rho)$ as $q$ in the equation above for succinctness. Here $\eps$ is small and $\theta \in [0,1]$ is large. This way, high-loss examples found in the semantic image manifold using simulation have a high probability of transferring to the real world. Since the simulator and real domain are different, this is a moderately strong assumption. Nevertheless, we show cases where this assumption holds in our experimental evaluations in Section~\ref{sec:method_weak}.

\paragraph{Finding Adversarial Parameters}
Our task is then to find $\rho$ such that the loss over samples generated with this latent variable is above the misclasification threshold $T$. One main difficulty in searching for latent variables that fulfill this condition is that in general the simulator $q$ is non-differentiable. Thus, we turn to black-box optimization methods to search for adversarial parameters. Specifically, we use a policy gradient method~\cite{reinforce}.

We define a policy $\pi_\omega$ parameterized by $\omega$ that can sample simulator parameters $\rho \sim \pi_\omega(\rho)$. We train this policy to generate simulator parameters that generate samples that obtain high loss when fed to the machine learning model $f$. For this we define a reward $R$ that is equal to the negative loss $L$ and we want to find the parameters $\omega$ that maximize $J(\omega) = \E_{\rho \sim \pi_{\omega}}[R]$. Following the REINFORCE rule we obtain gradients for updating $\omega$ as
\begin{equation}
  \nabla_{\omega} J(\omega) = \E_{\rho \sim \pi_\omega}\big[\nabla_{\omega} \log(\pi_{\omega}) R(\rho) \big] \;.
\end{equation}
An unbiased, empirical estimate of the above quantity is
\begin{equation}
  \mathcal{L}(\omega) = \frac{1}{K} \sum_{k=1}^{K} \nabla_{\omega} \log( \pi_{\omega})\hat{A}_k \;,
  \label{eq:policy_update}
\end{equation}
where $\hat{A}_k = R(\omega_k) - \beta$ is the advantage estimate, $\beta$ is a baseline, $K$ is the number of different parameters $\rho$ sampled in one policy forward pass and $R(\rho_k)$ designates the reward obtained by evaluating $f$ on $(x_k,y_k) \sim q(x_k, y_k|\rho_k)$. We show all of the steps of our method in \Algref{alg:main_alg} and we show an illustration of our method applied to the face verification scenario in Figure~\ref{fig:pipeline}.

\begin{algorithm}[]
  \SetAlgoLined
  \DontPrintSemicolon
  \KwResult{adversarial simulator parameters $\rho_k$ and adversarial sample $x_k$}
  \For{iteration=1,2,...}{
    Generate $K$ simulator parameters $\rho_k \sim \pi_{\omega}(\rho_k)$;
    Generate $K$ samples $(x_k,y_k) \sim q(x_k,y_k|\rho_k)$\;
    Test the discriminative model and obtain $K$ losses $L(f(x_k), y_k)$\;
    \If{$\exists k \in \{1, ..., K\} ; L(f(x_k), y_k) > T$}
    {Terminate and yield adversarial sample $x_k$ and adversarial simulator parameters $\rho_k$\;}
    Compute rewards $R(\rho_k)$\;
    Compute the advantage estimate $\hat{A}_k = R(\rho_k) - \beta$\;
    Update $\omega$ via~\eqref{eq:policy_update}\;
  }
  \caption{Our adversarial testing approach using a policy gradient method.}
  \label{alg:main_alg}
\end{algorithm}

\section{Finding Adversarial Regions}
Here we describe our method to find adversarial regions. Once an adversarial simulator latent vector $\rho_\text{adv} \in \mathbb{R}^n$ have been found using \Algref{alg:main_alg} we define a graph $G=(V,E)$. $V$ are the vertices of the graph, obtained by discretizing the space around the adversarial point in grid with spacing $\nu$ between vertices. The edges $E$ of the graph connect neighboring vectors, with each vector having $2n$ neighbors. We find the connected space of adversarial examples $\mathcal{R}_\text{adv}$ that is seeded by $\rho_\text{adv}$ by following \Algref{alg:region_alg}.

In essence, our method follows the general idea of an area flooding algorithm~\cite{coloring_book, tint_fill} with two main differences. First, that we discretize a continuous space that is $n$-dimensional instead of working on binary 2-dimensional image, and second, that we check for sample membership of $\mathcal{R}_\text{adv}$ by testing whether the model loss is higher than the adversarial threshold $L(f(x), y) > T$.

\begin{algorithm}[]
  \SetAlgoLined
  \DontPrintSemicolon
  \KwResult{connected space of adversarial examples $\mathcal{R}_\text{adv}$}
  \KwData{seed adversarial simulator parameters $\rho_\text{adv}$}
  $\mathcal{R}_\text{adv} = \{\rho_\text{adv}\}$\;
  Initialize a stack $\chi$.\;
  Push $2n$ neighbors of $\rho_\text{adv}$ to $\chi$.\;
  \For{$i$=1,2,...}{
    Pop $\rho_i$ from $\chi$\;
    Sample $(x_i,y_i) \sim q(x_i,y_i|\rho_i)$\;
    Test the discriminative model and obtain loss $L(f(x_k), y_k)$\;
    \If{$L(f(x_k), y_k) > T$}
    {$\mathcal{R}_\text{adv} = \mathcal{R}_\text{adv} \cup \{\rho_i\}$\;
    Push all neighbors of $\rho_i$ that have not been visited to $\chi$\;}
  }
  \caption{Finding connected spaces of adversarial examples.}
  \label{alg:region_alg}
\end{algorithm}

\section{Experimental Results}
\label{sec:exp}
\subsection{Controllable Face Simulation}
We use the FLAME face model~\cite{flame} as a controllable face simulator with the Basel texture model~\cite{basel}. FLAME uses a linear shape space trained from 3,800 3D scans of human heads and combines this linear shape space with an articulated jaw, neck, and eyeballs, pose-dependent corrective blendshapes, and additional global expression blendshapes. In this way, using shape and texture components we can generate faces with different identities. The synthetic faces that are generated in our work are new and do not mimic any existing person's features. By changing the pose and expression components we can add variability to these faces. Moreover, we have full control over the scene lighting and the head and camera pose and position. In order to render our scene we use the PyTorch3D rendering framework~\cite{ravi2020pytorch3d}. We extract the corresponding shape, texture and expression components from the real faces of the CASIA WebFace dataset using DECA~\cite{deca}.

\subsection{Models, Datasets and Infrastructure}
In our experiments we use the CASIA WebFace~\cite{yi2014learning} dataset for training the face recognition models and the LFW~\cite{LFWTech} dataset for real-world data testing. We use a Convolutional Block Attention Module (CBAM)~\cite{woo2018cbam} ResNet50 with the ArcFace~\cite{deng2019arcface} loss as our base face recognition model. We also test our method on MobileNet~\cite{howard2017mobilenets} and CBAM-Squeeze-Excitation-ResNet~\cite{hu2018squeeze} architectures and the CosFace~\cite{wang2018cosface} loss. We use a multivariate Gaussian policy $\pi(\rho) = \mathcal{N}(\mu_\pi, \sigma_\pi^2)$ where the variance is fixed $\sigma_\pi^2 = 0.05 \times I$ and $\mu_\pi$ is learned. For the random optimization baseline we use one Gaussian for each parameter type with standard deviation $\sigma_\text{rs} = \frac{w_p}{10} \times I$, where $w_p$ is the width of the parameter domain. For the Gaussian random sampling baseline we use a standard deviation $\sigma_\text{g} = \frac{w_p}{2}$. We use a GeForce RTX 2080 GPU with 11GB of memory to perform all of our experiments.

\subsection{Testing Weakened Models}
\label{sec:method_weak}
We present a way to verify that knowledge from simulated weaknesses translates to real-world weaknesses. We weaken two networks by training on the CASIA WebFace dataset with images that exhibit a yaw parameter $[-\infty,-0.5]$ and $[0.5,+\infty]$ filtered out. We extract the yaw parameter using DECA. We call these the \textit{Negative Yaw Filtered} (NYF) and \textit{Positive Yaw Filtered} (PYF) datasets/networks, respectively. Both datasets have roughly the same number of samples: the \textit{Negative Yaw Filtered} dataset has $\sim$ 440k training samples and the \textit{Positive Yaw Filtered} dataset has $\sim$ 449k samples. We also train a \textit{Normal} network on all of the $\sim$ 491k samples of the unfiltered CASIA WebFace dataset. We then test both the normal network and the yaw-weakened networks on simulated samples. We do this by generating two images of a same person, by fixing the shape, texture and expression parameters. The first image is a frontal image of the person. We vary the yaw component of the second image in the $[-1,1]$ range, where $-1$ and $1$ in the yaw component indicate a fully-profile face on the negative and positive sides, and compute the cosine similarity between the embeddings of the two images. This cosine similarity should be large given that the two images presented are of the same identity. A low cosine similarity means that the network has less confidence that the images show the same person. 

\begin{figure}
\centering
\includegraphics[width=0.7\columnwidth,trim={1cm 0cm 1.4cm 0cm}]{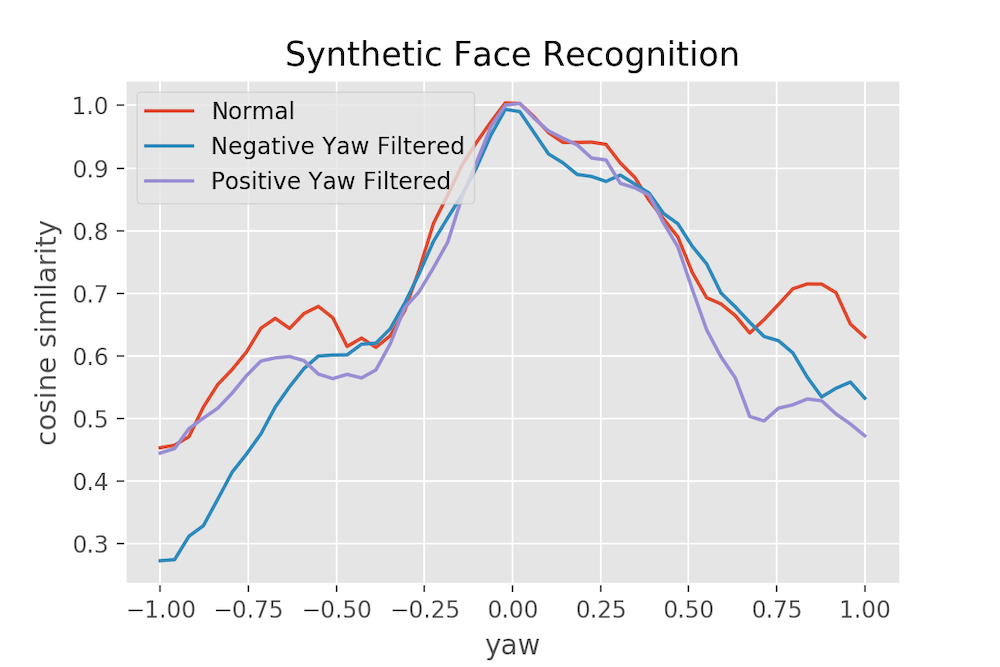}
\caption{Recognition cosine similarity between two simulated pairs (frontal and variable yaw) of the same identity (avg. over 25 different identities). The \textit{Negative Yaw Filtered} network exhibits less accurate predictions for highly negative yaw images than both the \textit{Positive Yaw Filtered} and \textit{Normal} networks. The \textit{Positive Yaw Filtered} network exhibits less accurate predictions for highly positive yaw images than both other networks.}
\vspace{-15pt}
\label{fig:yaw_and_adv_testing_runs}
\end{figure}

We plot this in Figure~\ref{fig:yaw_and_adv_testing_runs}, and observe that each yaw-weakened network makes less accurate predictions for images presenting high yaw in their respective weakness intervals. Note that all networks perform almost identically with frontal samples. Also, note that the normal network is almost always superior to the two weakened networks. This is a natural result of having 10\% more training data. This plot is an average over 25 different identities that we obtain by grid-sampling the first texture and shape components over the range $[-\sigma,\sigma]$.

We compute the area between the curves for the $[-1.0, -0.5]$, $[-0.5, 0.5]$ and $[0.5, 1.0]$ intervals. We observe in Table~\ref{table:yawchange_quantitative} (left) that in the $[-1.0, -0.5]$ yaw range, precisely where the NYF network has been weakened, the area between the Normal-NYF curves is large and the area between the Normal-PYF curves is small. Conversely, in the $[0.5, 1.0]$ range, where PYF has been weakened, we see that the difference between the Normal-PYF curves is large and the Normal-NYF difference is smaller. Also, we observe near identical differences between Normal-NYF and Normal-PYF in the $[-0.5, 0.5]$, which is a consequence of the lesser amount of training data of NYF and PYF networks. We also compute pairwise mean differences for the different populations of Normal, NYF and PYF networks and present them in Table~\ref{table:yawchange_quantitative} (right). We highlight in blue the statistically significant differences. We have similar results as in Table~\ref{table:yawchange_quantitative} (left).

This evidence indicates that when a weakness is purposefully created in a network by filtering out key samples in the real training dataset, we can retrieve this weakness using our face simulator. This gives credence to the idea that we are able to find simulated adversarial examples in the semantic image manifold that will give us knowledge about adversarial examples in the real world.

\begin{table*}[t]
\centering
\resizebox{0.8 \textwidth}{!}{
  \begin{tabular}{cccc}
    \toprule
     & \multicolumn{3}{c}{Area Between Curves}  \\
    $\downarrow$ Models / Yaw Interval $\rightarrow$ &  [-1.0, -0.5] & [-0.5, 0.5] & [0.5, 1.0]  \\
    \midrule
    Normal:NYF & \textbf{8.69} & 2.83 & 4.68 \\
    Normal:PYF & 2.71 & 2.76 & \textbf{8.46} \\
    \bottomrule
  \end{tabular}
  \hspace{30pt}
    \begin{tabular}{cccc}
    \toprule
     & \multicolumn{3}{c}{Mean Difference}  \\
    $\downarrow$ Models / Yaw $\rightarrow$  & -1.0 & 0.0 & 1.0  \\
    \midrule
    Normal-NYF & \textcolor{cobalt}{0.18} & \textcolor{cobalt}{0.01} & \textcolor{cobalt}{0.10} \\
    Normal-PYF & 0.01 & 0.00 & \textcolor{cobalt}{0.16} \\
    NYF-PYF & \textcolor{cobalt}{-0.17} & \textcolor{cobalt}{-0.01} & \textcolor{cobalt}{0.06} \\
    \bottomrule
  \end{tabular}
}
  \caption{Quantitative differences between evaluation of the purposefully weakened \textit{Negative Yaw Filtered} (NYF) and \textit{Positive Yaw Filtered} (PYF) and the \textit{Normal} on synthetic faces (bold values for emphasis). \textcolor{cobalt}{Blue values} in the table on the right mean the differences are statistically significant with $p<0.01$. 
  \label{table:yawchange_quantitative}}
  \vspace{-17pt}
\end{table*}

\subsection{Simulated Adversarial Testing of Face Recognition Models}
In this section we evaluate adversarial testing of face recognition models for face verification. Specifically, we generate samples using the FLAME face model and use our proposed search algorithm to fool face recognition models.

We train an ArcFace CBAM-ResNet50 on CASIA WebFace for 20 epochs. This network achieves a 99.1\% accuracy on the LFW test set for the face verification task. The evaluation task is face verification between two synthetic images of a same person's face, one frontal and one profile image. We vary the first 15 shape parameters as well as the first 15 texture parameters for our generated identities, ranging from $-2\sigma$ and $2\sigma$ where $\sigma$ is the standard deviation of each parameter in question.

\begin{figure}[t]
\centering
\includegraphics[clip,width=\columnwidth]{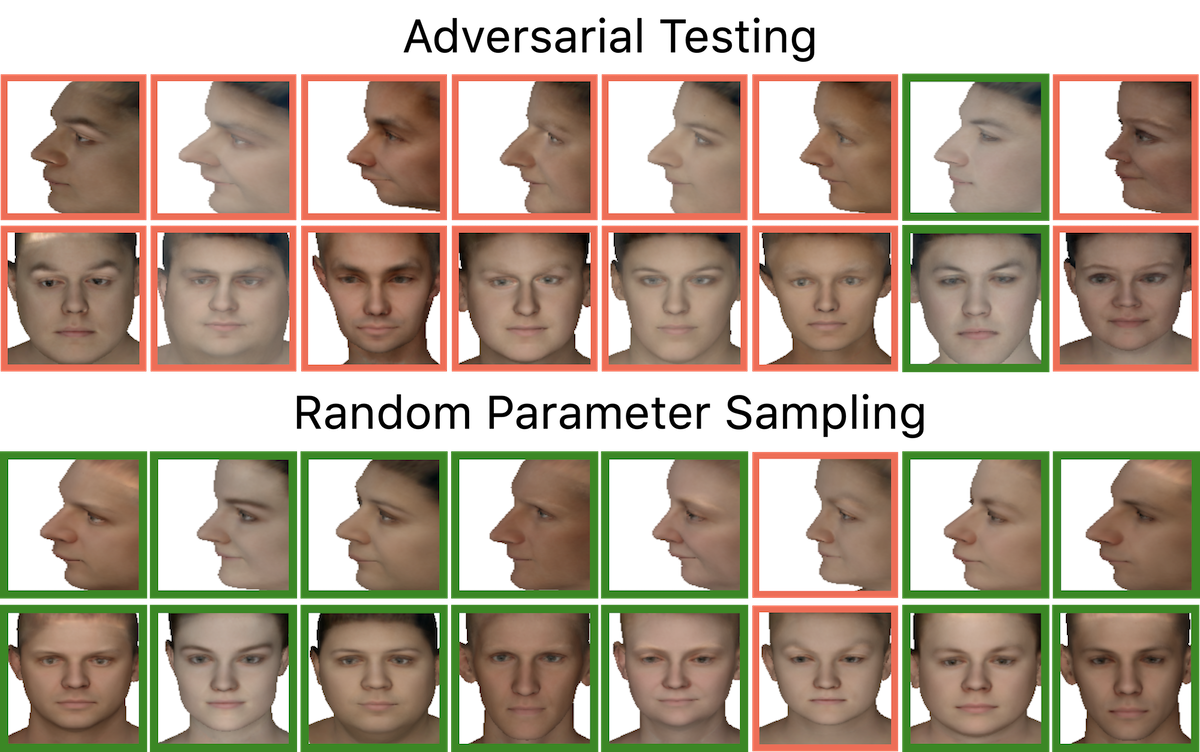}
\caption[]{Face models obtained using \textit{adversarial testing} (above) and random parameter sampling (below). A green border denotes pairs that are successfully verified as the same identity, whereas a red border denotes failed verification (model failure). We obtain adversarial samples using our \textit{adversarial testing} method more consistently than with random parameter sampling. Some recurring features of adversarial faces are ambiguous frontal/profile features (e.g. long nose, tucked jaw), pale/dark skin colors and left/right asymmetries.
\label{fig:samples_adv_testing}}
\vspace{-10pt}
\end{figure}

We propose testing the network using 100 identities obtained by random sampling these parameters following a uniform distribution. We also test the network using 100 runs of our adversarial testing algorithm (200 maximum iterations). In Table~\ref{table:adversarial_testing}, we show that the random sampling testing regime achieves an accuracy of 99\%, which is very close to the 99.1\% real-world accuracy of the network on the LFW test set. Using adversarial testing, the network exhibits an accuracy of 36\%, which is a marked drop in verification performance. We also compute the average cosine similarity between pairs, showing that adversarial testing generates highly adversarial samples (success threshold $T=0.298$) whereas random samples are highly non-adversarial on average. In Figure~\ref{fig:samples_adv_testing} we show a subset of the generated samples for both the adversarial testing (above) and random sampling (below). 

We perform further simulated adversarial testing experiments on several combinations of network backbones (CBAM-ResNet50, CBAM-SE-ResNet50~\cite{hu2018squeeze}, MobileNet) and face recognition losses (ArcFace, CosFace) trained on CASIA WebFace for 20 epochs. All networks achieve accuracies in the (98.85\%, 99.1\%) range on the LFW test set. We vary 30 shape parameters, 30 texture parameters ranging from $-2\sigma$ and $2\sigma$ where $\sigma$ is the standard deviation of each parameter. We also vary the yaw pose parameter within $[-1,+1]$, corresponding to variations of $[-\pi/2,+\pi/2]$ degrees and the pitch pose parameter from $[-1/4,+1/4]$ corresponding to variations within $[-\pi/8,+\pi/8]$. Thus, in this case our algorithm has to learn 62 parameters. This is a more challenging scenario due to the larger dimensionality of the policy output.

We perform 100 runs of our adversarial testing algorithm (200 maximum iterations), 100 runs of Random Optimization using a Gaussian sampling distribution and 1,000 iterations of uniform random sampling and Gaussian random sampling. We compare these testing methods in Table~\ref{table:adversarial_testing_2} and we show that the networks achieve very high accuracies for both random sampling regimes and for testing using random optimization. Using adversarial testing, all networks exhibit a marked drop in verification performance. There is also a large increase in the average cosine similarity between pairs, showing that adversarial testing generates highly adversarial samples (below success thresholds $T=(0.298, 0.237, 0.292, 0.294)$ respectively), whereas other methods generate ``easy'' samples on average. 

\begin{figure}[t]
\centering
\includegraphics[clip,width=0.65\columnwidth]{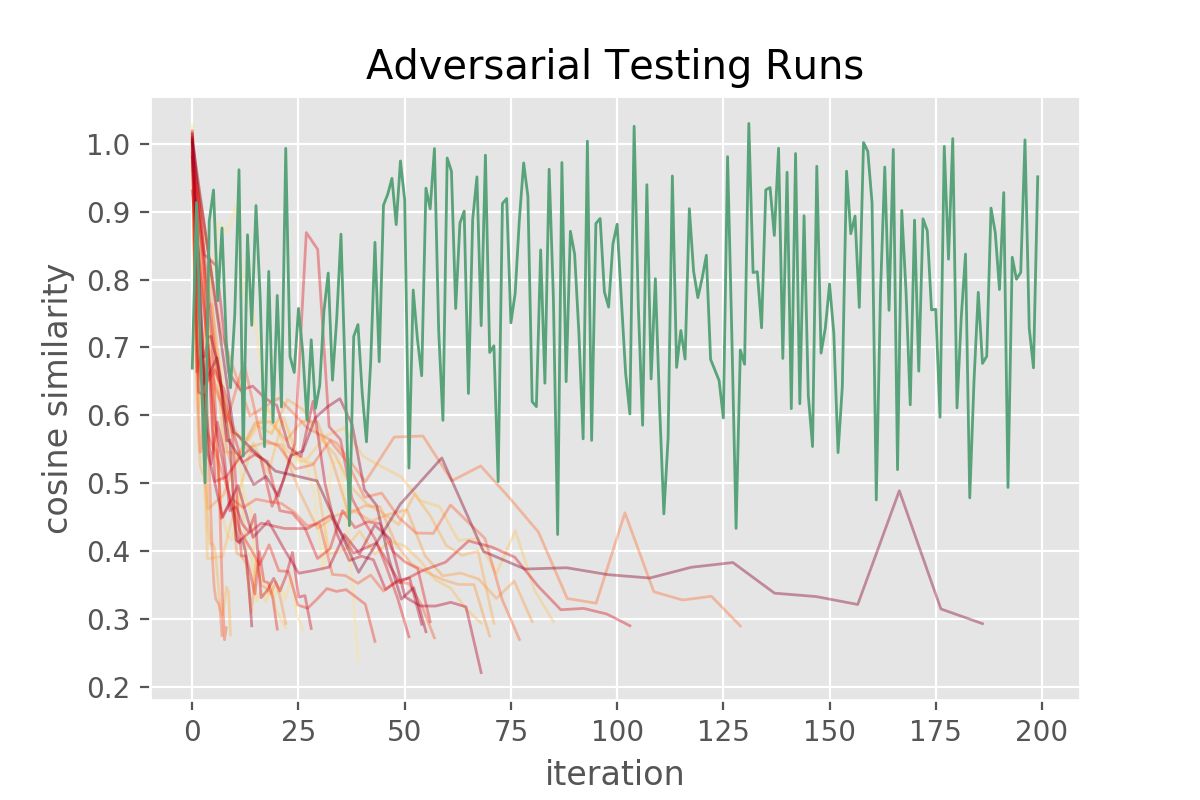}
\caption[]{
Cosine similarity for successful \textit{adversarial testing} (red) and random parameter sampling (green).}
\label{fig:adv_testing_supp_v2}
\vspace{-5pt}
\end{figure}

\begin{figure}[t]
\centering
\includegraphics[clip,width=0.9\columnwidth]{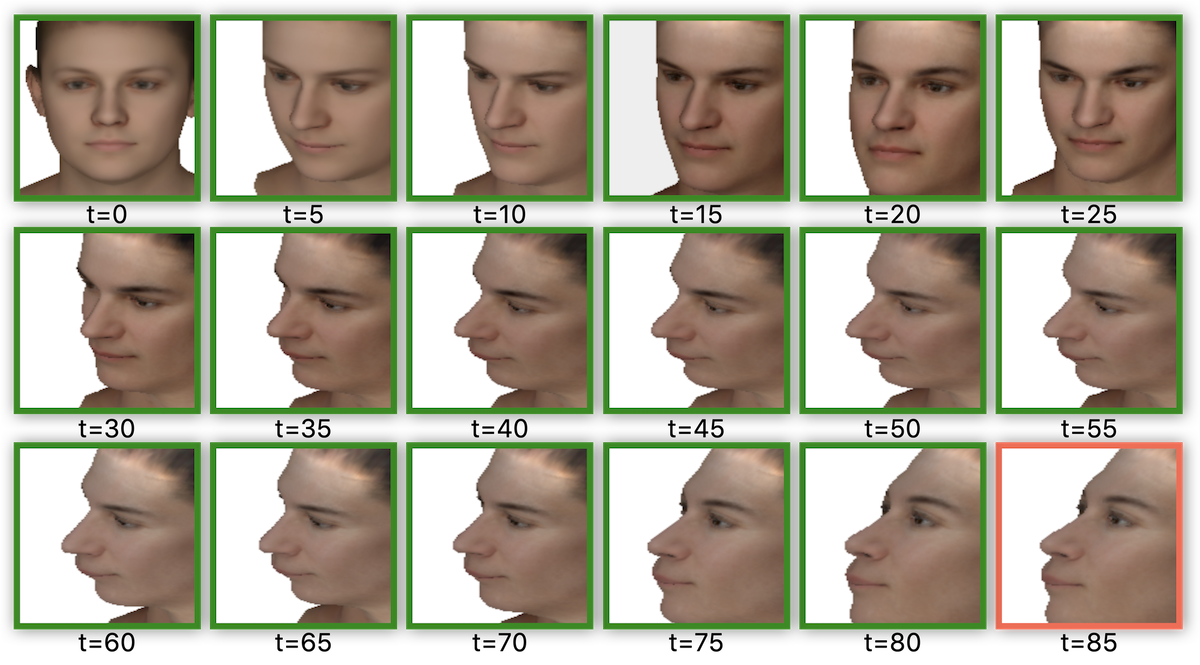}
\caption[]{
A sequence of generated synthetic samples undergoing \textit{adversarial testing} (left to right, top to bottom). Our method searches through all 30 shape, 30 texture and 2 pose parameters jointly to find an adversarial face. The border line colors denote whether the face recognition network can successfully verify the pairs, with red denoting a failed verification and green denoting a successful verification.}
\label{fig:samples_adv_testing_supp}
\vspace{-5pt}
\end{figure}

Further, for example, for ArcFace CBAM-ResNet50, adversarial testing achieves 51 adversarial samples over 12,587 iterations while random sampling achieves only one adversarial sample over 1,000 iterations. This makes adversarial testing 400\% more sample efficient than random sampling in this specific scenario. In some of our tested scenarios and depending on the number of iterations, random sampling was not able to find any adversarial samples. This is reflected by a 100\% face verification accuracy. In Figure~\ref{fig:adv_testing_supp_v2}, we show several successful adversarial testing runs (orange/red) and one random sampling run (green). Unsuccessful optimization attempts usually converge to low cosine similarity without becoming adversarial and remain in the high-dimensional local minima. Finally, we show an example of adversarial testing in action where all 30 shape, 30 texture and 2 pose parameters are being learned jointly in Figure~\ref{fig:samples_adv_testing_supp}. The algorithm finds an adversarial sample that reveals model weaknesses such as vulnerability to unusual poses, exaggerated facial features and distinct skin color.

\begin{table}[t]
\centering
  \caption{CBAM-ResNet50 face verification accuracy over synthetic datasets generated by uniform random sampling or by adversarial testing (Adv. Testing). We vary the identity by varying 15 shape parameters and 15 texture parameters.
  \label{table:adversarial_testing}}
\resizebox{\columnwidth}{!}{
  \begin{tabular}{cccc}
    \toprule
    Method &  Accuracy $\downarrow$ & Avg. Cosine Similarity $\downarrow$ \\
    \midrule
    Uniform Random & 99\% & 0.518 \\
    Adv. Testing & 36\% & 0.263 \\
    \bottomrule
  \end{tabular}
}
\end{table}

\begin{table*}[t]
\resizebox{\textwidth}{!}{
  \begin{tabular}{lcccccccc}
    \toprule
    & \multicolumn{2}{c}{\textbf{Uniform Random}} &
    \multicolumn{2}{c}{\textbf{Gaussian Random}} &
    \multicolumn{2}{c}{\textbf{Random Opt.}} &
    \multicolumn{2}{c}{\textbf{Adv. Testing}} \\
    Loss + Backbone & Acc. $\downarrow$ & Avg. CS $\downarrow$ & Acc. $\downarrow$ & Avg. CS $\downarrow$ & Acc. $\downarrow$ & Avg. CS $\downarrow$ & Acc. $\downarrow$ & Avg. CS $\downarrow$ \\
    \midrule
    ArcFace CBAM-ResNet50 & 99.9\% & 0.766 & 99.3\% & 0.695 & 93\% & 0.414 & 49\% & 0.282 \\
    CosFace CBAM-ResNet50 & 99.9\% & 0.696 & 99.6\% & 0.637 & 86\% & 0.318 & 57\% & 0.281 \\
    ArcFace SE-CBAM-ResNet50 & 99.8\% & 0.738 & 97.7\% & 0.663 & 73\% & 0.348 & 34\% & 0.305 \\
    ArcFace MobileNet & 100\% & 0.825 & 99.8\% & 0.751 & 96\% & 0.454 & 58\% & 0.372 \\
  \end{tabular}
}
  \caption{Comparison of different synthetic sampling techniques on different combinations of network backbones and face recognition losses. We vary 30 shape, 30 texture and 2 pose parameters.
  \label{table:adversarial_testing_2}}
  \vspace{-13pt}
\end{table*}

\subsection{Finding Adversarial Regions of Face Recognition Models}
We use our method described in \Algref{alg:region_alg} to find adversarial regions in the simulator latent space for face recognition models. We do this in the face verification scenario between a frontal image with neutral expression and a profile image with an open jaw. We vary the first shape and texture parameters to find an adversarial sample, and then find the connected spaces to those seed parameters. We also grid sample both parameters in order to plot the synthetic sample surface. We show the surface of all synthetic samples (blue), along with the adversarial region (red) and the adversarial threshold plane (orange) in Figure~\ref{fig:adv_regions}.

We are successful in finding the adversarial regions when they exist. We discover a surprising fact when plotting the synthetic loss landscape (Figure~\ref{fig:landscape_comparisons}) of all the tested networks. In this configuration with only 2 variable parameters, the only network with an adversarial region is ArcFace CBAM-ResNet50. Even though all networks have been trained in the same manner on the same dataset, the network backbone and the loss function change the loss landscape substantially. Some networks have a similar downward slope from negative shape towards positive shape, but some particularities arise in some. Strikingly, ArcFace MobileNet is the most robust of all the networks in this scenario with a landscape far above the misclassification threshold plane. The landscape shape is also completely different from the other networks.

\section{Related Work}

\begin{figure}[t]
\centering
\includegraphics[clip,width=0.49\columnwidth]{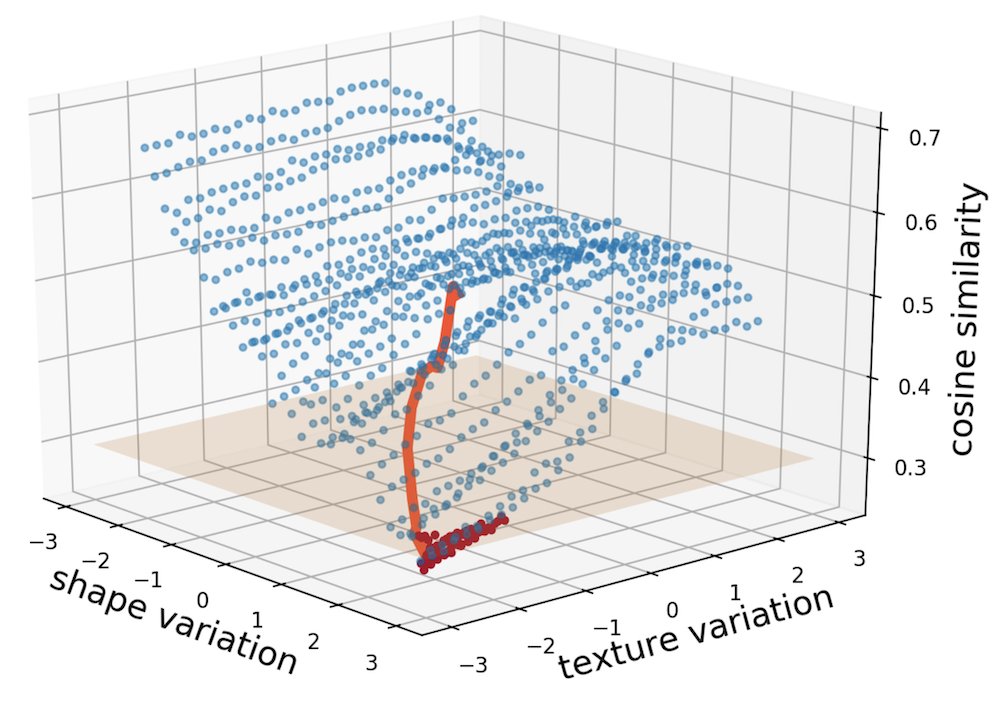}
\includegraphics[clip,width=0.49\columnwidth]{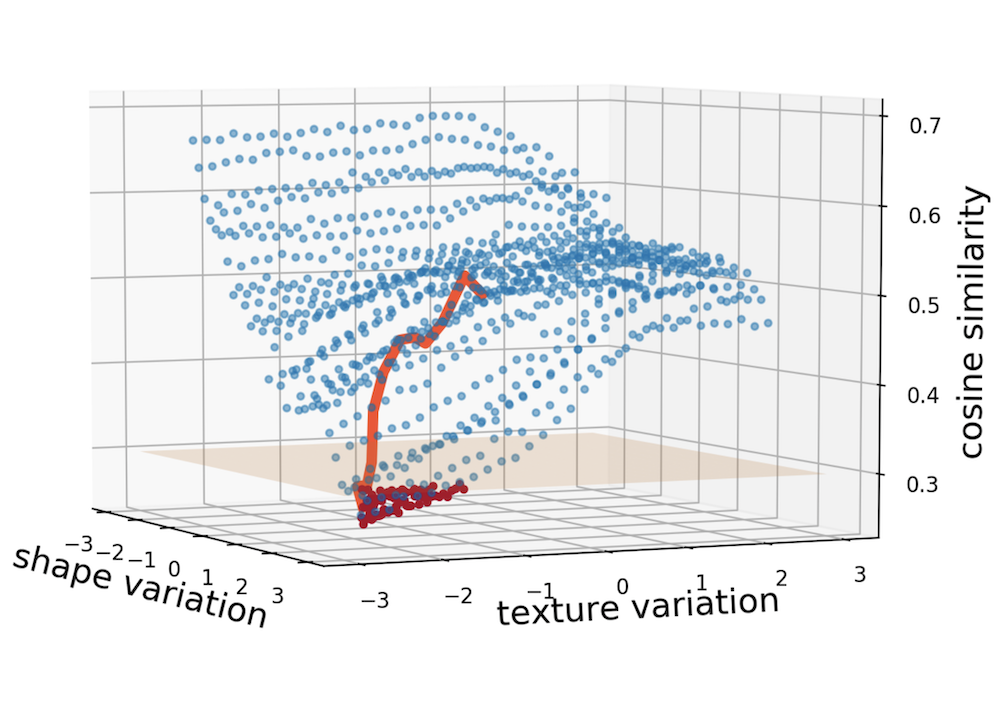}
\caption[]{Our algorithm finds the adversarial region (red) in the shape-texture landscape (blue). We plot the initial learning trajectory (lighter red) that yields the seed adversarial simulator parameters. We also plot the adversarial threshold plane (orange).
\label{fig:adv_regions}}
\vspace{-12pt}
\end{figure}

\begin{figure}[t]
\centering
\includegraphics[clip,width=0.8\columnwidth]{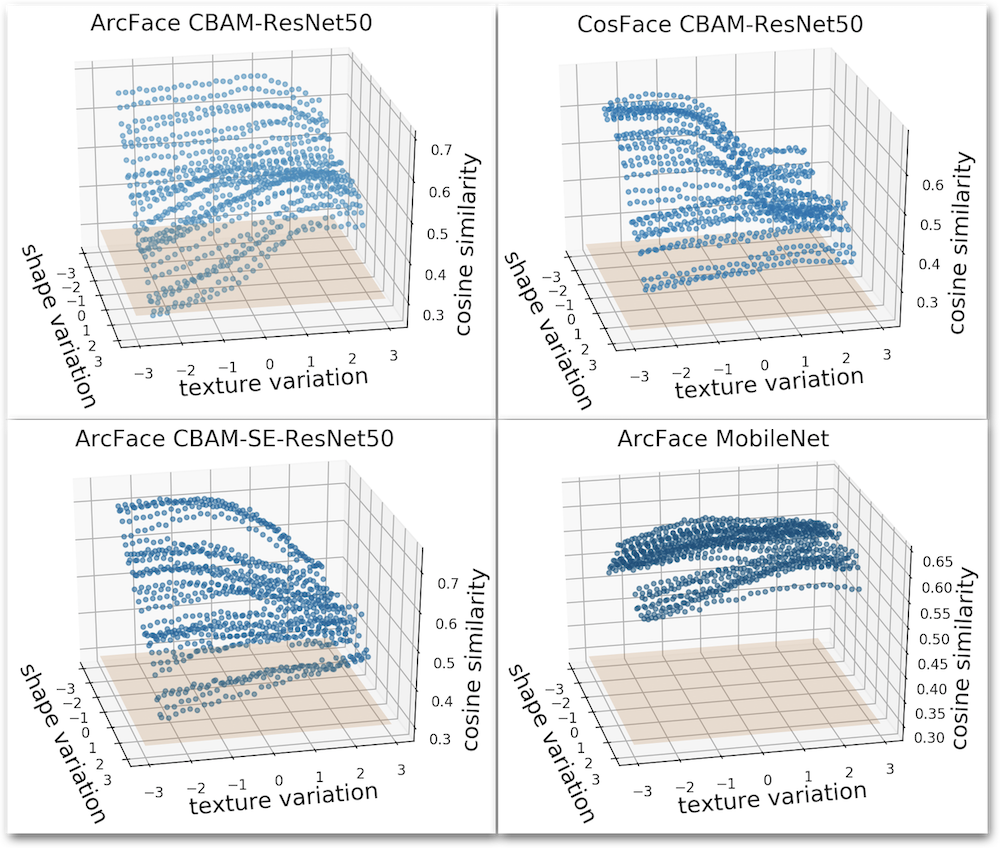}
\caption[]{Landscape comparisons for different network backbones and losses. Networks trained on CASIA WebFace.
\label{fig:landscape_comparisons}}
\vspace{-13pt}
\end{figure}

Testing computer vision models on synthetic data is not a new idea~\cite{pinto2008establishing, mayer2016large, johnson2017clevr, kortylewski2018empirically, kortylewski2019analyzing, ruiz2020morphgan}, although there is a relative paucity of work in this area. More common are investigations on training models on synthetic data~\cite{virtualkitti,synthia,playing_for_data,dosovitskiy2017carla,Ruiz_2018_CVPR_Workshops,kortylewski2018training,gecer2018semi,gecer2020synthesizing,marriott20213d}. Recent works even learn to adapt the generative distribution of synthetic data in order for the model to learn better representations~\cite{ruiz2018learning, louppe2017adversarial, Ganin2018SynthesizingPF, Beery_2020_WACV, Kar_2019_ICCV, andrychowicz2020learning} or adapt the pixels or features of the synthetic data to bridge the synthetic-to-real domain gap~\cite{Ganin:2016:DTN:2946645.2946704,Chen_2017_ICCV,tzeng2017adversarial, Tsai_2018_CVPR, hoffman2018cycada, peng2019moment}. In contrast to this body of work, we propose to search the parameter space of a simulator in order to test the model in an adversarial manner. There is very interesting work that adapts generative distributions in order to test models~\cite{alcorn2019strike, zeng2019adversarial, shu2020identifying}. In contrast to \cite{zeng2019adversarial, shu2020identifying} we test computer vision models that are trained on real data, which is a more challenging scenario since the domain shift problem has to be described and overcome. Different from \cite{zeng2019adversarial, alcorn2019strike, shu2020identifying} we work on the domain of face recognition instead of object classification or VQA, where we have a higher number of simulator parameters including shape, expression, texture, lighting and pose parameters. We search the parameter landscape using a continuous policy that explores all parameters simultaneously, which is important since model performance does not vary independently with each parameter (as Figure~\ref{fig:adv_regions} shows), and discrete changes in parameter space can yield high loss changes due to gradient sharpness. A final difference with these and work on traditional adversarial attacks~\cite{szegedy2013intriguing, papernot2017practical, carlini2017towards, explaining_adv, madry2018towards, ruiz2020disrupting, ruiz2020protecting} is that we present a method that not only finds one isolated adversarial example, but locates regions of them. There exist methods that propose objectives that locate regions of adversarial examples~\cite{salman2019provably}. In contrast, we explore the adversarial regions that lie in the latent space of a simulator instead of pixel space.

\section{Conclusion}
\label{sec:conclusion}
In this work we propose to test machine learning models by searching for semantically realistic adversarial examples using a simulator. We present a framework for simulated adversarial testing, as well as a method to find simulated adversarial examples. Finally, we present a method to find connected spaces of adversarial examples in the semantic space of latent variables and evaluate our methods on contemporary face recognition networks using a face simulator.

We find that face recognition networks that have real world weaknesses due to biased training sets with respect to pose can be analyzed using controllable simulated faces and these weaknesses can be discerned. We also find that contemporary face recognition networks are fooled by specific combinations of simulated face shapes and textures. Some recurring features of adversarial faces are ambiguous frontal/profile features (e.g. long nose, tucked jaw), pale/dark skin colors and left/right asymmetries. When such a network is tested using adversarial testing, it's accuracy plummets compared to random testing or testing on a real-world test set such as LFW. We show evidence that these adversarial examples are not isolated, but part of connected spaces of adversarial examples in the manifold of semantically plausible images. We also show that network loss landscapes can vary significantly depending on the network architecture and loss used, even though the training dataset is fixed. Even so, adversarial testing finds adversarial samples for all networks effectively. We will investigate this phenomenon in future work. Finally, we have an in-depth discussion of the limitations and potential negative impact of our work in the supplementary material.

\paragraph{Acknowledgments}
This work was supported in part by grants ONR N00014-21-1-2812 and NIH R01 EY029700 to Alan Yuille and a gift grant from Open Philanthropy to Cihang Xie.

{\small
\bibliographystyle{ieee_fullname}
\bibliography{main}
}

\clearpage

\section*{Supplementary Material: Simulated Adversarial Testing of Face Recognition Models}

\section*{Simulated Adversarial Testing Exploration}

\begin{figure*}[t]
\centering
\includegraphics[clip,width=\textwidth]{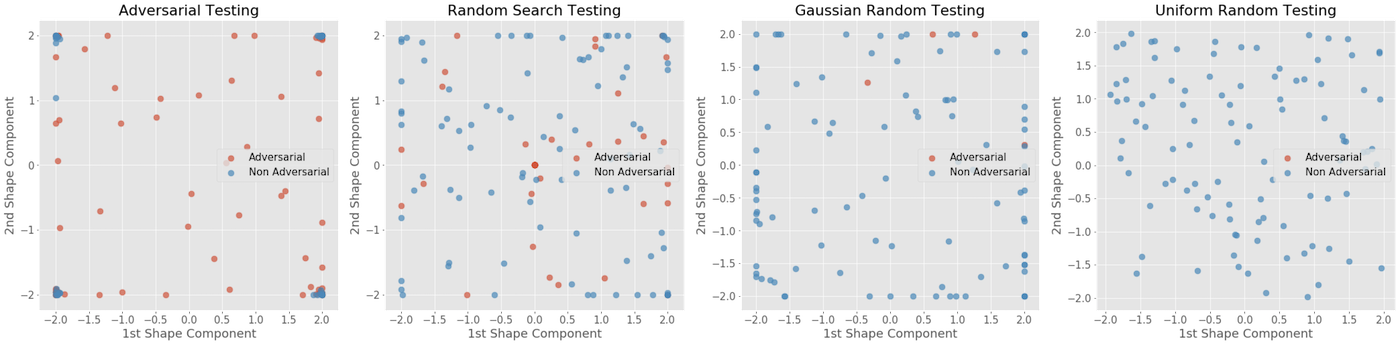}
\caption[]{Comparison of adversarial/non-adversarial samples for different testing methods, projected onto the 1st and 2nd shape component plane.
\label{fig:supp_12shape}}
\end{figure*}

In order to explore samples generated by simulated adversarial testing and other simulated testing techniques, we are able to project the shape and texture components of our samples onto a plane of two components. We do so for the first two shape components (roughly controlling for height and width of the face). We show the results for adversarial testing, random optimization, Gaussian random testing and uniform random testing in Figure~\ref{fig:supp_12shape}. First we observe the relative abundance of adversarial examples found using our method compared to other methods. Next, we can observe that adversarial testing not only finds adversarial examples, but that these examples are also very varied. We note that most unsuccessful runs of adversarial testing occur when the algorithm converges to local maxima that are located at the edges of the feasibility domain of $[-3,3]$. All plots are drawn from samples tested on an ArcFace IR-SE-CBAM-ResNet50.

We now show two plots for adversarial testing on this network where we project the samples on the plane generated by the 1st and 2nd shape components, and the 3rd and 4th shape components. Here we discover an interesting phenomenon. In the 1st-2nd shape component plane, samples are varied and seem roughly uniformly distributed in the space. This means that although the 1st and 2nd shape component clearly have a role in finding adversarial samples, adversarial samples can be found with many different 1st and 2nd shape component values. The second plot shows that for the 3rd and 4th shape components, our adversarial sampling method clearly favors/disfavors some pockets in the space. For example we find that samples with average 3rd and 4th shape components tend not to be found by adversarial testing. We can test the hypothesis of whether these values are non adversarial by higher-dimensional grid search, although this would be time consuming. Another idea is to limit the feasibility domain to these average 3rd and 4th shape components, and run many instances of adversarial testing. If few or no adversarial samples are found then there is a chance that this is a space that is highly non-adversarial. We believe these types of projections can give a strong intuition over what features affect network performance. In Figure~\ref{fig:supp_flame_shape_front} and \ref{fig:supp_flame_shape_side} we show the first four shape component variations for the FLAME model in frontal and profile poses. We can see that the 3rd and 4th shape component variations, while not overly noticeable in the frontal pose, introduce features that are only visible in the profile pose. For example when the 4th shape component is varied in the positive direction it tucks the subject's jaw in. This introduces a frontal/profile ambiguity, and the face verification network has a harder time correctly verifying pairs of these faces since it takes as input both the frontal and profile face images. Similarly, the 3rd shape component introduces a protuberance of the subject's head when varied in the positive direction, which is much more apparent in the profile image. This is congruent to the adversarial faces that we obtain in Figure 3 of the main paper that show frontal/profile ambiguous features.

\begin{figure}[t]
\centering
\includegraphics[clip,width=\columnwidth]{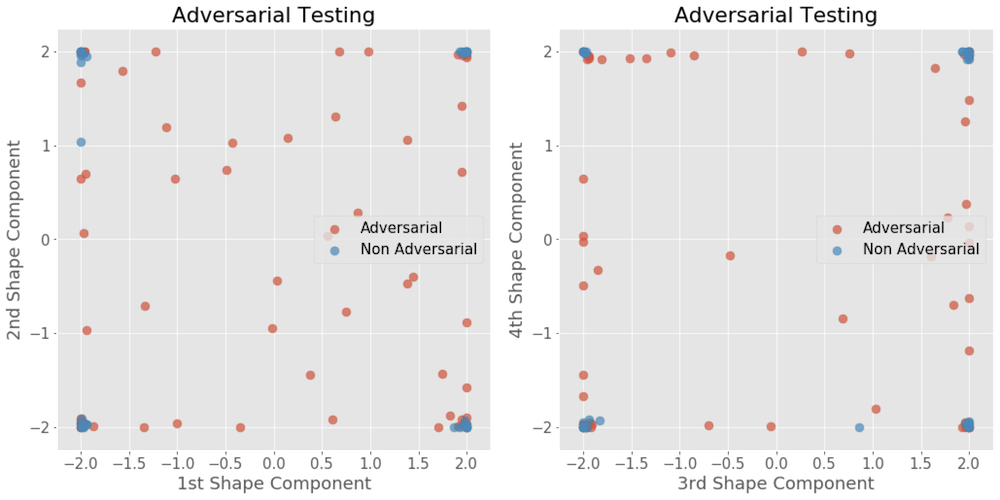}
\caption[]{Comparison of adversarial/non-adversarial samples for simulated adversarial testing, projected onto the 1st/2nd and 3rd/4th shape component planes.
\label{fig:supp_shape_comparison}}
\end{figure}

\begin{figure}[t]
\centering
\includegraphics[clip,width=\columnwidth]{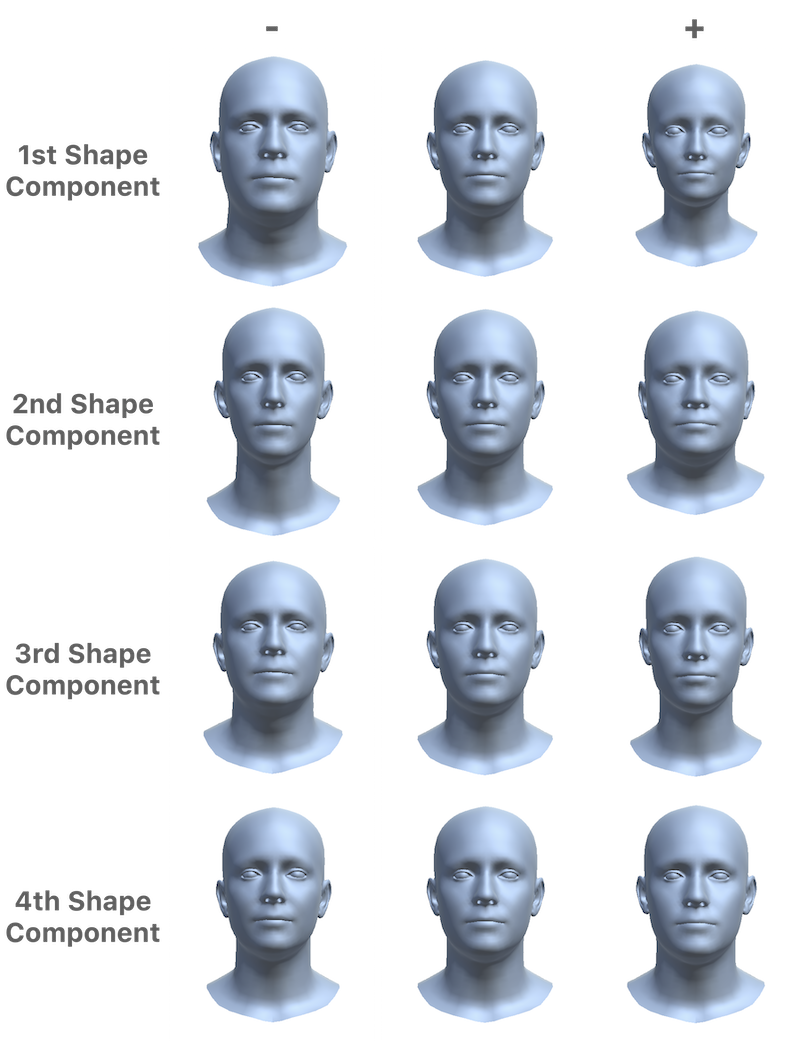}
\caption[]{Shape variations along different shape components for the FLAME model in a frontal pose.
\label{fig:supp_flame_shape_front}}
\end{figure}

\begin{figure}[t]
\centering
\includegraphics[clip,width=\columnwidth]{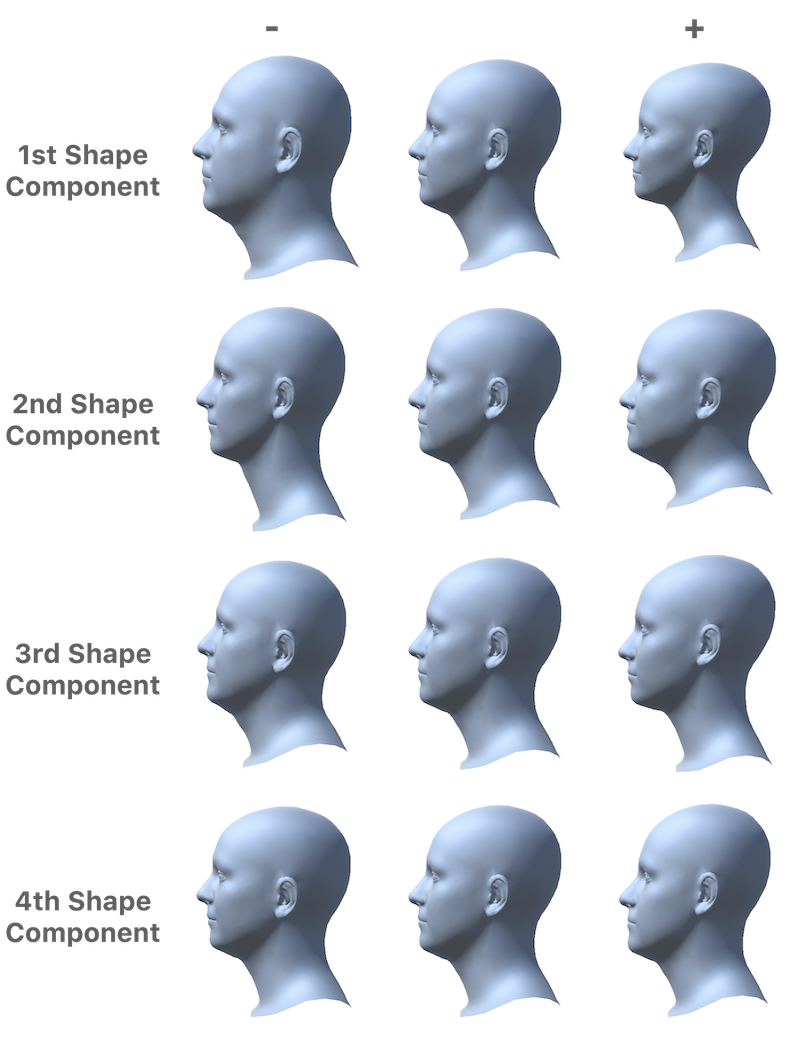}
\caption[]{Shape variations along different shape components for the FLAME model in a profile pose.
\label{fig:supp_flame_shape_side}}
\end{figure}

\section*{Limitations and Future Work}

Even though our adversarial testing algorithm using reinforcement learning is much more effective than random optimization and other sampling methods, it does not have a perfect rate of finding adversarial examples. It sometimes converges to local maxima that are hard to classify but nonetheless non-adversarial. We believe that one of the weaknesses is that it can get stuck in the boundary limits for the parameters that are being varied, and it has a hard time getting out of that space. This is especially a large problem in higher dimensional spaces. We will investigate this weakness in future work.

While we do exhibit for the first time the fact that two face recognition networks trained on the same data with different architectures and losses have vastly different loss landscapes when face shape and texture are varied, and thus are learning different things, we have not yet found suitable hypotheses that are verified by data that explain \textit{why} this is the case. We believe this phenomenon requires more in-depth research and are working on verifying some of our hypotheses for our next work.

One of the major limitations of our work is the fact that we find adversarial examples that are simulated. We believe that the most challenging aspect of this research direction is verification of simulated adversarial samples using real data. This aspect is so challenging that it has been neglected by prior research. Nevertheless, many works have been accepted to top conferences without treating this specific problem due to the potential they have in furthering our knowledge about robustness issues of neural networks with realistic variation of stimuli~\cite{alcorn2019strike, kortylewski2018empirically, shu2020identifying, ruiz2020morphgan, zeng2019adversarial}. Even the best simulators that are currently available to the computer vision research community exhibit a substantial domain gap with real data~\cite{shah2018airsim, smplx, dosovitskiy2017carla, flame}. For this reason, it is difficult to verify the transfer capability of certain features. Of the different attributes that were available to us, head pose is one of the most reliable in terms of transfer due to several reasons: the ability to easily extract it from real images using a head pose estimation network, the 1-to-1 correspondence between head pose in simulated and real situations, and the low-dimensional nature of the attribute that can be more easily analyzed and plotted in a curve as shown in Figure 2. Due to all of these reasons we present the first link between simulated adversarial samples in the simulated and real world using this attribute. In the near-future, with more advanced simulators, we expect work to be able to confirm many more strong links between simulated and real samples. Just as \cite{alcorn2019strike} found that camera pose influences the predictions of a neural network in simulated data, we show that pose, shape and texture jointly influence predictions of a face recognition network, but we go one step further and show that pose similarly impacts performance in the real and simulated world. Finally, in principle, it is almost impossible to find a real face that is arbitrarily close to any face we simulate. This is simply due to the fact that shape and texture are very high-dimensional, such that a point has very few close neighbors given a fixed-size real dataset. To find a very close sample we would have to collect a dataset that is extremely large.

\section*{Societal Impact}
The plausible negative social consequences of this work are tightly linked with overall negative consequences of facial analysis systems. An approach that improves testing for face recognition systems such as the one we propose can be used to improve recognition rates on minorities, persecuted groups and oppressed individuals. This is a larger problem acting on any work that can potentially impact facial analysis, and we argue that our work has an asymmetric potential for applications that have a positive social impact. Given that researchers have proven that there exists gender and racial bias of beneficial face analysis systems~\cite{buolamwini2018gender, Garcia_2019, grother2019face}, by better testing such systems these biases can be diagnosed and mitigated, meaning that minorities can more readily benefit from these technologies. 

Another important point is that a major bottleneck for our work is a simulator that is expressive and realistic. Bias and lack of expressiveness of a simulator might mean that bias in the face recognition network is not correctly detected. We urge developers of future simulators to take into account the bias of their training population in order to increase the expressiveness of their simulator and decrease the bias. We also urge them to understand the power of such a tool for robustness and bias analysis and to distribute the model responsibly, similar to the FLAME head model~\cite{flame} team.

\end{document}